\documentclass[conference]{IEEEtran}

\usepackage{amsfonts,amsmath,amsthm}
\usepackage{graphicx,xspace,epsfig,syntonly,dsfont}
\usepackage{url,cite,bm,bbm}
\usepackage{algorithmicx}
\usepackage{balance}
\usepackage{subfigure}
\usepackage{stfloats}
\usepackage{diagbox}
\usepackage{enumitem}

\long\def\symbolfootnote[#1]#2{\begingroup
\def\thefootnote{\fnsymbol{footnote}}
\footnote[#1]{#2}\endgroup}

\IEEEoverridecommandlockouts

\allowdisplaybreaks[4]

\usepackage{xcolor}
\begin{document}

\title{Attention-based Neural Load Forecasting: \\ A Dynamic Feature Selection Approach}

\author{
    \IEEEauthorblockN{Jing Xiong\IEEEauthorrefmark{1}, Pengyang Zhou\IEEEauthorrefmark{2}, Alan Chen\IEEEauthorrefmark{3} and Yu Zhang\IEEEauthorrefmark{4}}
    \IEEEauthorblockA{\IEEEauthorrefmark{1}Dept of Electrical and Computer Engineering, UC Santa Cruz.\, Email: \texttt{jxiong20@outlook.com}}
    \IEEEauthorblockA{\IEEEauthorrefmark{2}Dept of Computer Science and Engineering, UC Santa Cruz.\, Email: \texttt{pengyangzhou@outlook.com}}
    \IEEEauthorblockA{\IEEEauthorrefmark{3}Santa Cruz High School.\, Email:  \texttt{alanphchen@gmail.com}}    
    \IEEEauthorblockA{\IEEEauthorrefmark{4}Dept of Electrical and Computer Engineering, UC Santa Cruz.\, Email: \texttt{zhangy@ucsc.edu}}
    \thanks{This work was supported in part by a Seed Fund Award from CITRIS and the Banatao Institute at the University of California, and the Hellman Fellowship.}
}

\maketitle

\begin{abstract}
Encoder-decoder-based recurrent neural network (RNN) has made significant progress in sequence-to-sequence learning tasks such as machine translation and conversational models. Recent works have shown the advantage of this type of network in dealing with various time series forecasting tasks. The present paper focuses on the problem of multi-horizon short-term load forecasting, which plays a key role in the power system's planning and operation. Leveraging the encoder-decoder RNN, we develop an attention model to select the relevant features and similar temporal information adaptively. First, input features are assigned with different weights by a feature selection attention layer, while the updated historical features are encoded by a bi-directional long short-term memory (BiLSTM) layer. Then, a decoder with hierarchical temporal attention enables a similar day selection, which re-evaluates the importance of historical information at each time step. Numerical results tested on the dataset of the global energy forecasting competition 2014 show that our proposed model significantly outperforms some existing forecasting schemes.
\end{abstract}

\section{Introduction}\label{sect:intro}
An accurate forecast of electricity load demand plays a crucial role in reliable and efficient planning and operation of power grids \cite{HONG2016896}. Based on forecast horizon, short, medium and long term load forecasts are provided by system operators and utilities for various applications and business needs. The ahead-of-time horizon of short-term load forecasting (STLF) typically ranges from several hours to a few days. STLF is indispensable for day-ahead unit commitment, market clearing, setting generation reserve, energy bidding, as well as real-time power dispatch \cite{feng2019reinforced}.

There are three different types of approaches for STLF: (i) time series analysis, (ii) classical machine learning algorithms, and (iii) deep learning models. Time series analysis, such as autoregressive moving average (ARMA), requires a meticulous preprocessing to make a time series stationary \cite{contreras2003arima}. Moreover, time series approaches are sensitive to irrelevant features and may fail to capture a long term dependency. Classical machine learning models such as support vector regression (SVR) and random forest can avoid those drawbacks. They are more tolerant of irrelevant features and more powerful in capturing the nonlinear behavior of electricity load. However, for multi-horizon load forecasting, a predetermined nonlinear model may prevent such an approach from learning the true underlying nonlinearity effectively \cite{qin2017dual}.

Over the past decade, deep neural networks (DNNs) have shown their powerful capability to learn very complicated input-output relationships in various fields including natural language processing \cite{young2018recent} and computer vision \cite{voulodimos2018deep}. Among different structures of neural networks, recurrent neural networks (RNNs) have been widely used in the sequence-to-sequence (seq2seq) learning tasks. They have shown the efficacy for time series forecasting \cite{cao2012forecasting}. However, the so-called gradient vanish is one of the major limitations for a plain RNN when dealing with long-term dependencies \cite{bengio1994learning}. To address such an issue, long short-term memory networks (LSTM) were proposed \cite{hochreiter1997long}. Three gates including the input gate, forget gate, and output gate are designed to control information flows. Through these gates, the relevant information is kept for long term memory while the other information is forced to be ignored.

Most seq2seq models leverage the encoder-decoder structure \cite{vaswani2017attention,cho2014properties}. A sequence of input features is first encoded as a single fixed-length vector by the encoder, based on which the decoder yields the output. However, this may make it difficult for the encoder-decoder approach to cope with long input sequence when the neural network is supposed to be able to compress all the necessary information.
To bypass this limitation, the attention mechanism was developed to search for a set of positions in a source where the most relevant information is concentrated \cite{bahdanau2014neural}.
For this new paradigm, a context vector is built to bridge the gap between the encoder and the decoder, which is filtered for each output time step.

Recently, various deep learning techniques have been utilized for STLF. A framework of deep residue networks is introduced in \cite{chen2018short}, which is able to integrate domain knowledge and build blocks based on the understanding of the task. A single bi-directional LSTM (BiLSTM) layer is implemented based on attention and rolling update \cite{wang2019bi}. However, the single BiLSTM layer lacks the ability to fully translate the input sequence to the target output, especially for multi-horizon forecasting. Compared with classical machine learning models, the development of deep learning for load forecasting is still premature.

Feature selection plays a crucial role in machine learning and deep learning \cite{kwak2002input}. Traditional feature selection methods, such as filter, wrapper, and embedded methods, are often implemented separately from the main learning algorithms \cite{stanczyk2015feature, feng2017data}. Qin \emph{et~al} \cite{qin2017dual} proposed an input attention layer that can be trained simultaneously with the model. However, this feature selection scheme is based only on the previous information, and hence cannot capture all the information of the entire input sequence.

In this paper, we propose an attention-based neural load forecasting (ANLF) framework, which aims to improve the forecasting accuracy by dynamically selecting the relevant features and similar temporal information. The contribution of our work is three-fold: (i) attention-based dynamic feature selection is introduced for STLF, which can weigh each feature based on the entire input sequence; (ii) hierarchical temporal attention layer is built to incorporate similar day information; and (iii) an end-to-end ANLF model is developed that significantly outperforms traditional methods.

 \section{Proposed Approach}\label{sec:approach}
\begin{figure*}[!tb]
\centering
  \includegraphics[width=0.95\textwidth]{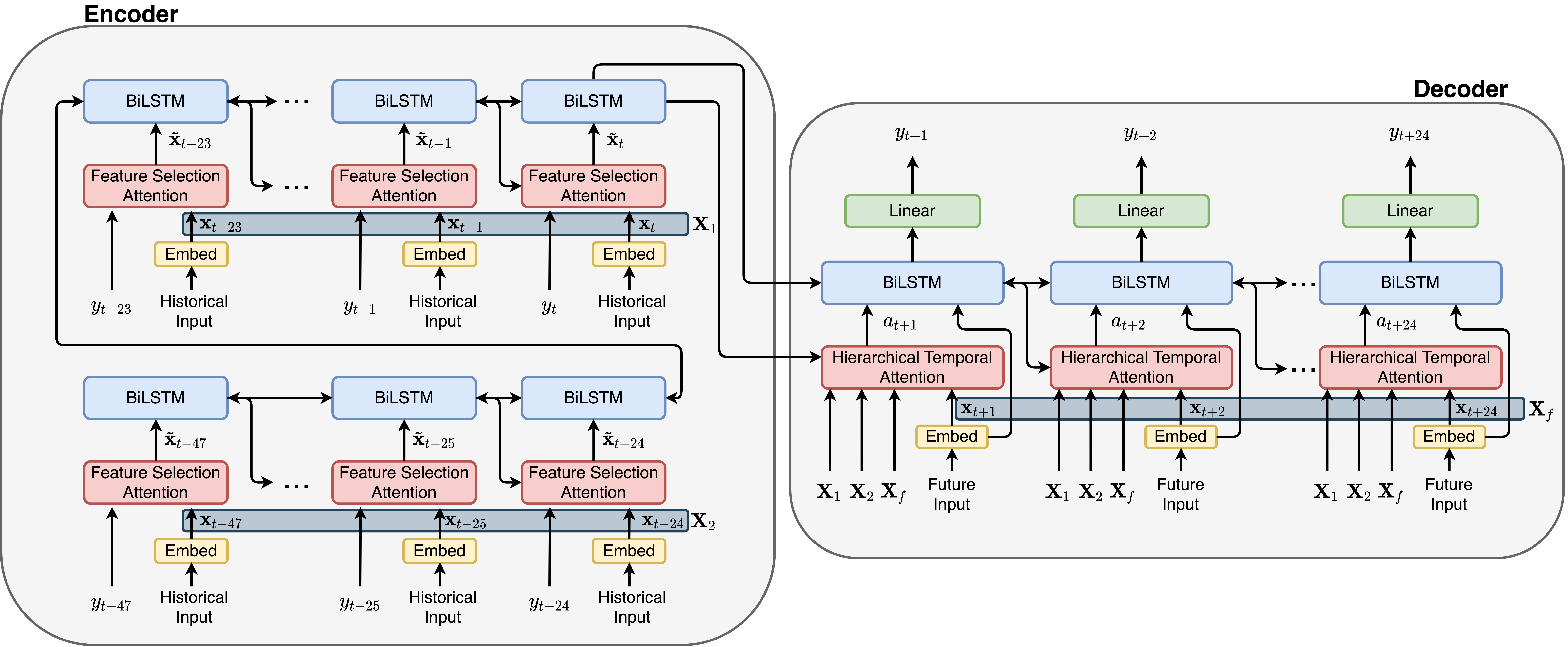}
  \caption{The network architecture of the proposed model. 
  The feature selection attention in the encoder is designed to adaptively weigh the different input features. In contrast, the hierarchical temporal attention in the decoder focuses more on the temporal similarity to incorporate similar day information. 
 To simplify the demonstration, the structure in the figure only reflects the scenario of forecasting the next day based on the previous two days. This can be easily extended to a general case.}
  \label{fig:Parnn_structure}
  \vspace{-0.5cm}
\end{figure*}

The novel architecture of our proposed attention-based DNN model, whose backbone is a competitive encoder-decoder structure, is illustrated in Fig.~\ref{fig:Parnn_structure}. In this section, we will elaborate the details of each module by highlighting the new design of dynamic feature selection and similar day selection for multi-horizon STLF.

\subsection{Input Feature Embedding and Problem Statement}\label{s}

By transforming categorical data into a numeric vector, embedding is widely used in machine learning when inputs contain one or more discrete values or items from a finite set of choices. For STLF, the inputs can contain meteorological conditions (e.g., temperature, humidity, wind speed and direction, etc), indicator of holidays, and utility discount programs. In this work, we simply use the one-hot encoding for categorical features. After embedding, the inputs of the encoder are the concatenation of embedded categorical features, continuous numeric features, and historical target values (active power demand) at each time step $t\in [t-T_h+1,t]$, while inputs of the decoder are the concatenation of embedded categorical features, continuous numeric features at each time step $t\in [t+1,t+T_f]$, and historical feature matrix $\mathbf{X}_i$ and future feature matrix $\mathbf{X}_f$, where $T_h$ and $T_f$ are the window size of historical and future data, respectively.

Let $\mathbf{X}=(\mathbf{x}^1,\mathbf{x}^2,\dots,\mathbf{x}^n) \in \mathbb{R}^{(T_h+T_f)\times n}$ 
collect all embedded input features across the entire horizon of interest. Thus, each row vector of $\mathbf{X}$ denoted as $\mathbf{x}_{t} = (x_{t}^1,x_{t}^2,\dots,x_{t}^n)$ represents all $n$ input features at time $t$. For the multi-horizon STLF task, given the historical feature inputs $\mathcal{X}_h :=\{\mathbf{x}_{h}\}_{h=t}^{t-T_h+1}$ and the corresponding target values $\mathbf{y}_h=(y_{t},y_{t-1},\dots,y_{t-T_h+1})^{\top}$, we aim to learn a function $f(\cdot)$ that can predict future target values $\mathbf{y}_f=(y_{t+1},y_{t+2},\dots,y_{t+T_f})^{\top}$ for the given future features $\mathcal{X}_f := \{\mathbf{x}_{f}\}_{f=t+1}^{t+T_f}$; i.e., $\mathbf{y}_f=f(\mathcal{X}_f~|~\mathbf{y}_h,\mathcal{X}_h)$.

\subsection{Encoder and Decoder}\label{secBin}

The encoder-decoder structure is a workhorse in the state-of-the-art deep neural networks. For time series forecasting, the encoder maps historical input features $\mathbf{x}_{t}$ and output $y_{h,t}$ at each time step $t$ to a hidden vector $\mathbf{h}_t$ that is passed to the decoder. Then, the decoder uses the last step hidden state of the encoder as its initial hidden state and outputs future target values based on future feature inputs.

\subsubsection{Encoder with feature selection attention}
As shown in Fig.~\ref{fig:Parnn_structure}, the encoder consists of two sublayers: a feature selection attention layer and a BiLSTM layer. The time range for the encoder is from time $t-T_h+1$ to $t$.

Feature selection is key for almost all machine learning tasks. Irrelevant features can significantly affect the model performance. By using the feature selection attention, the proposed model is able to adaptively weigh different features and give more attention to features that contribute more to the target values. The weight $\alpha_t^k$ of each feature $k$ at time $t$ is calculated via the softmax operator
\begin{equation} \label{eq:weightAlpha}
    \alpha_{t}^{k}=\frac{\exp \left(e_{t}^{k}\right)}{\sum_{i=1}^{n} \exp \left(e_{t}^{i}\right)},\, k = 1,2,\ldots, n
\end{equation}
where $e_t^k$ is the $k$-th element of the vector $\mathbf{e}_t = (e_t^1,e_t^2,\dots,e_t^n)^{\top} \in \mathbb{R}^n$ that is given by
\begin{equation} \label{eq:featureAttention_e}
    \mathbf{e}_{t}=\mathbf{V}_e\tanh\left(\mathbf{W}_{e}\left[{\mathbf{h}_{t-1}^e}^{\!\!\top} ; \mathbf{x}_t; y_t\right]^{\top}\right).
\end{equation}
The weight matrices $\mathbf{V}_e \in \mathbb{R}^{n\times p}$ and $\mathbf{W}_{e} \in \mathbb{R}^{p\times (2 hs+n+1)}$ are trained jointly with the proposed model. $p$ is a hyper-parameter while $hs$ is the hidden size of the BiLSTM. $[\mathbf{a}; \mathbf{b}]$ denotes the concatenation of two row vectors $\mathbf{a}$ and $\mathbf{b}$. $\mathbf{h}_{t-1}^e$ is the hidden vector of the encoder BiLSTM at time $t-1$. We omit the bias term for succinctness. Then, the weighted feature input is given by 
$
\tilde{\mathbf{x}}_t=\left(\alpha_{t}^{1} x_{t}^{1}, \alpha_{t}^{2} x_{t}^{2}, \ldots, \alpha_{t}^{n} x_{t}^{n}\right)
$.

The BiLSTM consists of two LSTM layers in opposite directions which can capture the complete information of the entire input sequence at each time step. Let $ \mathbf{h}_{t}^f$,  $ \mathbf{h}_{t}^b \in \mathbb{R}^{hs}$ denote the hidden state of forward and backward LSTM at time $t$, respectively. Given the weighted input $\tilde{\mathbf{x}}_t$, the hidden states of the encoder are updated iteratively from time $t-T_h+1$ to $t$ as follows \cite{fan2019multi}
\begin{subequations} \label{eq:biLSTM}
\begin{align}
    \mathbf{h}_{t}^f &=\mathrm{LSTM}^f(\mathbf{h}_{t-1}^f,\Tilde{\mathbf{x}}_t)\vspace{1ex}\\
    \mathbf{h}_{t}^b &=\mathrm{LSTM}^b(\mathbf{h}_{t+1}^b,\Tilde{\mathbf{x}}_t)\vspace{1ex}\\    
    \mathbf{h}_{t}^e &=\left[{\mathbf{h}_{t}^f}^{\top};{\mathbf{h}_{t}^b}^{\top}\right]^{\top}\vspace{1ex}.
\end{align}
\end{subequations}
The $\mathrm{LSTM}(\cdot)$ involves the following detailed updates \cite{hochreiter1997long}:
\begin{subequations}\label{eq:LSTM}
\begin{align}
\mathbf{i}_t &=\sigma\left(\mathbf{W}_{ix} \tilde{\mathbf{x}}_t^{\top}+\mathbf{b}_{ix}+\mathbf{W}_{ih} \mathbf{h}_{t-1}+\mathbf{b}_{ih}\right) \\
\mathbf{f}_t &=\sigma\left(\mathbf{W}_{fx} \tilde{\mathbf{x}}_t^{\top}+\mathbf{b}_{fx}+\mathbf{W}_{fh} \mathbf{h}_{t-1}+\mathbf{b}_{fh}\right) \\
\mathbf{g}_t &=\tanh \left(\mathbf{W}_{gx} \tilde{\mathbf{x}}_t^{\top}+\mathbf{b}_{gx}+\mathbf{W}_{gh} \mathbf{h}_{t-1}+\mathbf{b}_{gh}\right) \\
\mathbf{o}_t &=\sigma\left(\mathbf{W}_{ox} \tilde{\mathbf{x}}_t^{\top}+\mathbf{b}_{ox}+\mathbf{W}_{oh} \mathbf{h}_{t-1}+\mathbf{b}_{oh}\right) \\
\mathbf{c}_t &=\mathbf{f}_t \odot \mathbf{c}_{t-1}+\mathbf{i}_t \odot \mathbf{g}_t \\
\mathbf{h}_t &=\mathbf{o}_t \odot \tanh \left(\mathbf{c}_t\right),
\end{align}
\end{subequations}
where $\mathbf{i}_t$, $\mathbf{f}_t$, $\mathbf{g}_t$, $\mathbf{o}_t$ are the input, forget, cell, and output gates, respectively. $\mathbf{W}_{\cdot,x} \in \mathbb{R}^{hs\times n}$, $\mathbf{W}_{\cdot,h} \in \mathbb{R}^{hs\times hs}$ are the weight matrices; $\mathbf{b}_{\cdot,\cdot} \in \mathbb{R}^{hs}$ the bias vectors; $\mathbf{c}_t\in \mathbb{R}^{hs}$the cell state; $\sigma(\cdot)$ the sigmoid function; and $\odot$ the Hadamard product (element-wise multiplication). 

\subsubsection{Decoder with hierarchical temporal attention}
The decoder leverages the encoder's last-step hidden state and cell state as its initial state. Using the last forecast value as the input for the next time step may introduce error propagation. Instead, we design a BiLSTM along with a feed-forward layer to predict the final target sequence once the BiLSTM learns all input information. The time range for the decoder is from time $t+1$ to $t+T_f$.

Incorporating the information of similar days and hours has been considered in the literature for load forecasting; see e.g., \cite{feng2018hourly, barman2018regional}. However, such  information is often treated as additional input features or used to generate separate models. In this paper, we develop a novel hierarchical temporal attention layer, which incorporates a similar day soft selection to re-evaluate the importance of historical information at each time step $t$.

Consider using previous $M$ days of historical data to forecast the hourly loads for the next day, where each day includes $t_d =24$ data points. Thus, we have $T_h = M\times t_d$ and $T_f = t_d$. 
Let $\mathbf{X}_i=(\mathbf{x}_i^1,\mathbf{x}_i^2,\dots,\mathbf{x}_i^n)\in \mathbb{R}^{t_d\times n}$ and $\mathbf{X}_f=(\mathbf{x}_f^1,\mathbf{x}_f^2,\dots,\mathbf{x}_f^n)\in \mathbb{R}^{t_d\times n}$ collect the historical features for day $i$ and the future features for the next day, respectively. We use the sum of feature-by-feature dissimilarities $D(\mathbf{X}_i,\mathbf{X}_f) = \sum_{k=1}^n \|\mathbf{x}_i^k-\mathbf{x}_f^k\|_2$ to quantify the distance between all features of those two days. Then, the similar day weight $\gamma_{i}$ is calculated as the softmax of the reciprocal of the distance:
\begin{equation} \label{eq:weightGamma}
    \gamma_{i}=\frac{\exp\left(D^{-1}(\mathbf{X}_i,\mathbf{X}_f)\right)}{\sum_{i=1}^M\exp\left(D^{-1}(\mathbf{X}_i,\mathbf{X}_f)\right)},\, i=1,2,\ldots,M.
\end{equation}

When forecasting load at time $t$, not all historical data contribute equally to the model's output. Hence, the attention mechanism facilitates the extraction of historical information that is more important to the current forecast value. Let subscript $i$ denote the $i$-th day and $j$ for $j$-th hour. Then, the attention weight $\beta_{i,j,t}$ is given by
\begin{equation} \label{eq:weightBeta}
    \beta_{i,j,t}=\frac{\exp(d_{i,j,t})}{\sum_{i=1}^{M}\sum_{j=1}^{t_d}\exp(d_{i,j,t})},
\end{equation}
where $d_{i,j,t}$ is the $(i\times t_d + j)$-th element of vector $\mathbf{d}_t = (d_t^1,d_t^2,\dots,d_t^{T_h})^{\top} \in \mathbb{R}^{T_h}$, which is given as
\begin{equation} \label{eq:temporalAttention_d}
    \mathbf{d}_{t}=\mathbf{V}_d\tanh\left(\mathbf{W}_{d}\left[{\mathbf{h}_{t-1}^d}^{\top} ; \mathbf{x}_t\right]^{\top}\right).
\end{equation}
The two weight matrices $\mathbf{V}_d \in \mathbb{R}^{T_h\times p'}$ and $\mathbf{W}_{d} \in \mathbb{R}^{p'\times (2hs+n)}$ are trained jointly with the proposed model. 
$\mathbf{h}_{t-1}^d$ is the hidden vector of the decoder BiLSTM at time $t-1$.

To this end, let $\mathbf{h}_{i,j}$ denote the historical hidden state for the $j$-th hour in the $i$-th day  from the encoder. The context vector of hierarchical temporal attention is calculated as
$\mathbf{a}_{t}=\sum_{i=1}^{M}\sum_{j=1}^{t_d} \gamma_{i}  \beta_{i,j,t}\mathbf{h}_{i,j}
$.
Given the future feature input $\mathbf{x}_t$ and context vector of hierarchical temporal attention $\mathbf{a}_t$, the hidden state of the decoder is updated iteratively from time $t+1$ to $ t+T_f$ as $\mathbf{h}_{t}^d=\mathrm{BiLSTM}(\mathbf{h}_{t-1}^d,\mathbf{h}_{t+1}^d,[\mathbf{x}_t;\mathbf{a}_t^{\top}])$.

Finally, a fully connected layer with the rectified linear unit (ReLU) activation function is used to transform the hidden information to the forecast output
\begin{equation} \label{eq:output}
    \mathbf{y}_f=\mathbf{V}_y\text{ReLU}(\mathbf{W}_y\mathbf{h}_{t+1:t+T_f}^d),
\end{equation}
where $\mathbf{V}_y \in \mathbb{R}^{T_f\times p''}$ and $\mathbf{W}_y \in \mathbb{R}^{p''\times (T_f \times 2hs)}$ are weight matrices; and vector $\mathbf{h}_{t+1:t+T_f}^d = \left[ {\mathbf{h}_{t+1}^d}^{\top}; {\mathbf{h}_{t+2}^d}^{\top}; \dots; {\mathbf{h}_{t+T_f}^d}^{\top}  \right]^{\top}$. In the decoder module, both $p'$ and $p''$ are hyper-parameters to be tuned.

\section{Numerical Results} \label{sec:tests}

\subsection{Data Description}
We evaluate the proposed model on an hourly load data introduced by the Global Energy Forecasting Competition 2014 extended (GEFCom2014-E) dataset \cite{HONG2016896}. The dataset includes load demand, temperature, and timestamps across a total of 9 years between January 1, 2006 to December 31, 2014. We use the data from the year 2010 to 2012 for training, the year 2013 for validation, and the year 2014 for testing. 

The data of the past seven days are used to forecast the next day. Specifically, the inputs of the model are listed in Table \ref{table:1}. The dataset is first standardized (subtract the mean and divide by the standard deviation) to mitigate the magnitude impact of different input features.

\begin{table}[t]
\centering
\caption{Inputs of the proposed model.}
\label{table:1}

\renewcommand{\arraystretch}{1.2}
\resizebox{0.5\textwidth}{!}
{%
\begin{tabular}{lll}
\hline
Input & Size & Description \\ \hline
$\mathbf{y}_h$ & $168\times 1$ & Historical target values  \\
Temperature & $192\times 1$  & Historical and future temperature \\
Holiday  & $192\times 1$ & Holiday and non-holiday indicators \\ 
Hour of Day  & $192\times 24$ & One-hot encoding \\
Day of Week  & $192\times 7$ & One-hot encoding   \\ 
Month of Year  & $192\times 12$ & One-hot encoding \\ \hline
\end{tabular}%
}
\end{table}

\subsection{Baseline and Model Setup}

To show the effectiveness of our proposed model, we compare two different types of models: classical machine learning models and LSTM based models. We implement the Support Vector Regression (SVR), Random Forest (RF), and Gradient Boosting Machine(GBM) by using Scikit-Learn 0.23.2. For LSTM based methods, we consider the basic Encoder-Decoder based LSTM (EDLSTM), Encoder-Decoder based BiLSTM (EDBiLSTM) using bidirectional LSTM in the Decoder, proposed method with attention only applied in the encoder (eAttention), proposed method with attention only applied in the decoder (dAttention) and proposed method with feature weight calculated by Random Forest \cite{saeys2008robust} (RFAttention). All the LSTM based methods are trained by the Adam optimizer with the mean squared error (MSE) loss function, which are implemented via PyTorch 1.6.0. To ensure fair comparisons, all models share the same training, validation, and testing datasets as well as input features. The grid search is used to tune the hyper-parameters. The final setting are listed as follows:
\begin{itemize}[leftmargin=*]
  \item SVR: linear kernel with $C=0.1$.
  \item RF: number of estimators $=1,000$, max-depth $=20$, min-samples-split $=2$, min-samples-leaf $=1$.
  \item GBM: loss $=$ ``ls'', learning rate $=0.01$, number of estimators $=1,000$, max-depth $=5$, min-samples-split $=2$, min-samples-leaf $=15$.
  \item EDLSTM: batch $=64$, hidden size $=1,024$, epochs $=5$.
  \item EDBiLSTM: batch $=64$, hidden size $=1,024$, epochs $=5$.
  \item RFAttention: batch $=128$, hidden size $=256$, epochs $=15$.
  \item dAttention: batch $=256$, hidden size $=128$, epochs $=5$.
  \item eAttention: batch $=256$, hidden size $=128$, epochs $=20$.
  \item ANLF: batch $=128$, hidden size $=256$, epochs $=5$.
\end{itemize}

\subsection{Numerical Results}

\begin{figure*}[!th]
\centering
  \includegraphics[scale = 0.05]{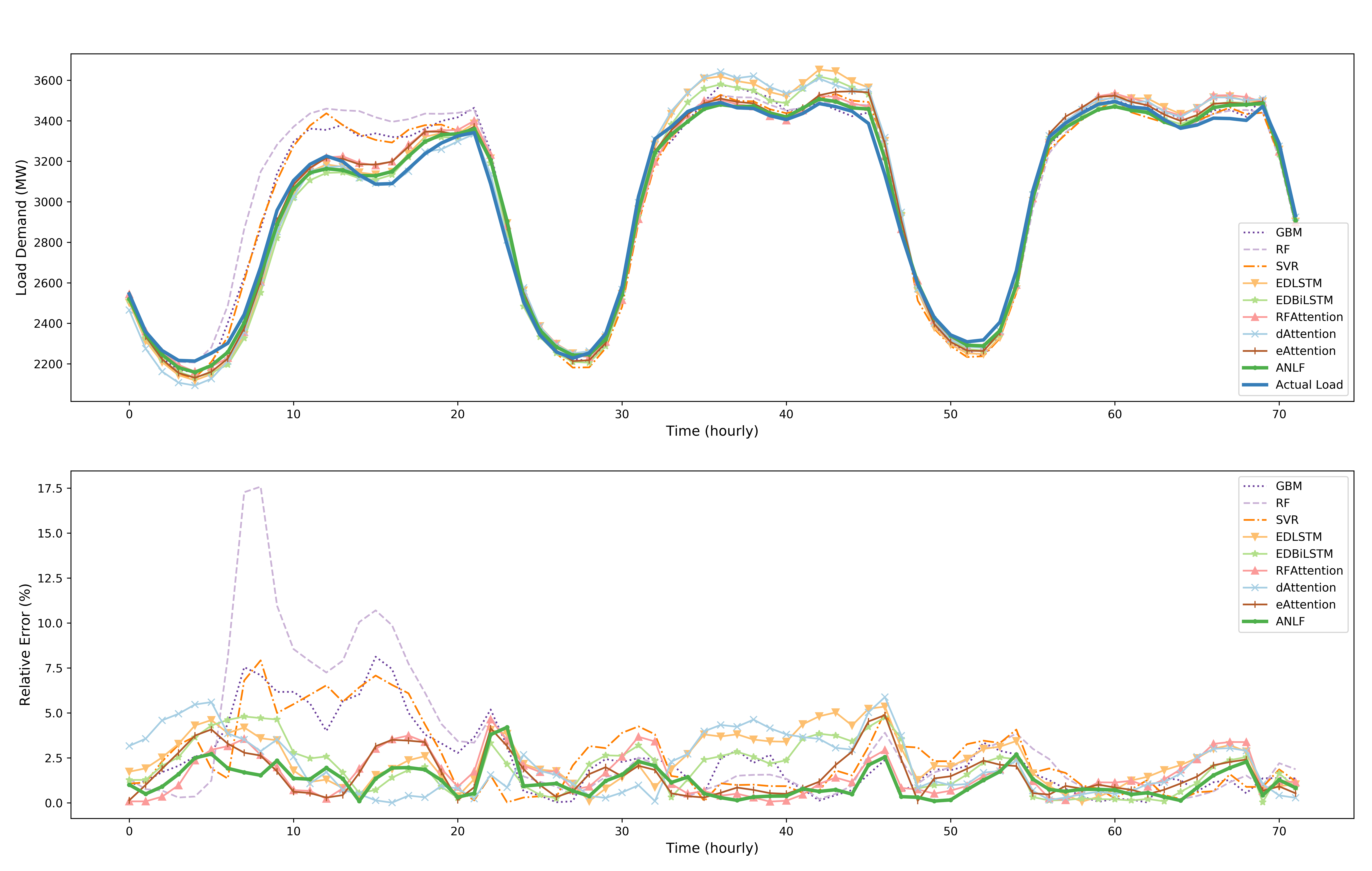}
  \caption{The detailed forecasting performance of nine different models over three days.}\label{fig:result_detail}
  \vspace{-0.5cm}
\end{figure*}

Table~\ref{table:overall_result} summarizes the overall performance of all nine competing models across the entire testing horizon. To evaluate the forecasting performance, four error metrics are considered including mean absolute error (MAE), root mean square error (RMSE), mean absolute percentage error (MAPE), and normalized root mean square error (nRMSE)~\cite{shcherbakov2013survey}.  GBM is the best method among those classical machine learning models, which achieves similar performance as dAttention and EDBiLSTM. In LSTM-based methods, we can see a clear increase of accuracy by adding the attention mechanism to both the encoder and the decoder. Results for RFattention and ANLF further approve the effectiveness of the proposed feature selection layer.  Clearly, our proposed ANLF approach significantly outperforms all other baseline models. For the STLF task, the simulation results corroborate the merit of integrating the attention mechanism to focus on more relevant features and similar temporal references. 
 A detailed forecasting performance over 3 days is shown in Fig.~\ref{fig:result_detail}. The relative error of between the forecast value $\hat{y}$ and the true value $y$ is defined as $\mathrm{RE}=\frac{|y-\hat{y}|}{y}\times 100\%$.

\begin{table}[t]
\centering
\caption{Forecasting Errors over the test year 2014.}
\label{table:overall_result}
\renewcommand{\arraystretch}{1.2}
\begin{tabular}{ccccc}
\hline
	Model & MAE & RMSE & MAPE (\%) & nRMSE (\%) \\ \hline
	GBM & 71.48 & 94.61 & 2.15 & 2.84 \\ 
	RF & 85.36 & 113.67 & 2.55 & 3.42 \\ 
	SVR & 79.23 & 108.27 & 2.33 & 3.26 \\ 
	EDLSTM & 74.73 & 97.95 & 2.26 & 2.94 \\ 
	EDBiLSTM & 71.48 & 92.94 & 2.17 & 2.79 \\ 
	RFAttention & 73.72 & 97.83 & 2.23 & 2.95 \\ 
	dAttention & 70.91 & 91.42 & 2.15 & 2.75 \\ 
	eAttention & 65.33 & 87.85 & 1.97 & 2.65 \\ 
	ANLF & \textbf{64.80} & \textbf{86.74} & \textbf{1.93} & \textbf{2.61} \\\hline
\end{tabular}
\end{table}

\section{Conclusion}
In this paper, we develop an end-to-end attention-based neural load forecasting framework for multi-horizon STLF. For the proposed approach, the dynamic feature selection layer combined with a BiLSTM encoder is designed to extract more relevant features. Then, a BiLSTM decoder with a hierarchical temporal attention layer decodes the next day load based on the future input features. The temporal attention layer provides a systemic way to incorporate similar day information. The extensive simulation results show the effectiveness of our proposed approach that has an edge over the state-of-the-art benchmarks. 

\nocite{*}
\bibliographystyle{IEEEtran}
\bibliography{GM2021,IEEEabrv}

\end{document}